\pdfoutput=1
\documentclass[11pt]{article}
\usepackage{acl}
\usepackage{times}
\usepackage{latexsym}
\usepackage[T1]{fontenc}
\usepackage[utf8]{inputenc}
\usepackage{microtype}
\usepackage{inconsolata}
\usepackage{amsfonts,amsmath,amssymb,nicefrac,bbm}
\usepackage{graphicx,booktabs,multirow,makecell,subcaption,xcolor}
\usepackage{algorithm,algpseudocode}
\usepackage{enumitem}
\usepackage{tikz,pgfplots}
\pgfplotsset{compat=1.18}
\usetikzlibrary{arrows.meta,positioning,fit,shapes,backgrounds,calc}
\usepackage{tcolorbox,mdframed}
\usepackage{tabularx}
\usepackage{stfloats}
\usepackage{xspace}
\usepackage{amsthm}

\newcommand{\mage}{\textsc{Mage}\xspace}
\newcommand{\gepa}{\textsc{Gepa}\xspace}

\newcommand{\memmod}{\mathcal{M}}
\newcommand{\dataset}{\mathcal{D}}
\newcommand{\prompt}{\pi}
\newcommand{\prompts}{\Pi}
\newcommand{\trace}{\tau}
\newcommand{\RR}{\mathbb{R}}
\newcommand{\eg}{\textit{e.g.,}\xspace}

\definecolor{mageblue}{RGB}{31,119,180}
\definecolor{mageorange}{RGB}{214,90,48}
\definecolor{magegreen}{RGB}{15,110,86}
\definecolor{magepurple}{RGB}{83,74,183}
\definecolor{magegray}{RGB}{136,135,128}
\definecolor{mageamber}{RGB}{186,117,23}
\definecolor{promptbg}{RGB}{245,247,250}
\definecolor{promptborder}{RGB}{180,190,210}

\newmdenv[backgroundcolor=promptbg,linecolor=promptborder,linewidth=0.8pt,
  innerleftmargin=8pt,innerrightmargin=8pt,innertopmargin=6pt,
  innerbottommargin=6pt,skipabove=4pt,skipbelow=4pt,roundcorner=4pt]{promptbox}

\title{\mage: Understanding Stability--Performance Trade-offs\\
in Multi-component Prompt Optimization}

\author{Prateek Singh \\
  \texttt{prateek29singh@gmail.com}}

\begin{document}
\maketitle

%% ─────────────────────────────────────────────────────────────────
\begin{abstract}
%% ─────────────────────────────────────────────────────────────────

How do different components of iterative prompt optimization
interact---and what happens when they are combined?

We investigate this question through \mage{} (Memory-Augmented
Goal-directed Prompt Evolution), a controlled \textit{analysis
framework} for studying component interaction in prompt optimization.
\mage{} is not proposed as a superior optimizer in absolute terms;
it integrates three mechanisms---episodic memory, multi-objective
Pareto selection, and adaptive evaluation---as a platform for
controlled ablation.
Our experiments uncover a previously unreported phenomenon, the
\textit{Prompt Optimization Coupling Effect} (POCE): when multiple
stochastic optimization signals operate within a closed reflective
loop, they interact in ways that simultaneously improve performance
and amplify variance---behavior that cannot be predicted by analyzing
components in isolation.

Three main findings emerge.
First, \textit{failure-grounded reflection is essential}: methods that
rely only on scores (OPRO) or abstract critique (Self-Refine) fail to
improve prompts, either remaining stuck at the seed or degrading
performance.
Second, \mage achieves $\mathbf{46.4\%}$ versus \gepa's $34.0\%$ on
GSM8K-Hard ($+12.4\%$, $P(\text{\mage}{>}\text{\gepa}){=}0.998$, 5 seeds
on gpt-4o-mini), with comparable variance ($\pm7.3\%$ vs.\ $\pm7.0\%$).
Third, increasing candidate diversity reveals the clearest POCE signal:
expanding the candidate pool from $n{=}3$ to $n{=}5$ improves mean
accuracy by $+21.6\%$ while increasing variance by $3.7\times$.

We further validate on Llama~3.1~8B and show POCE is
\textit{headroom-dependent}: when the base model already achieves high
accuracy, variance amplification disappears.
In an on-device deployment validation using a 1.5B model,
\mage converges in $2.0{\pm}0.0$ iterations across 5 seeds
(100\% convergence rate), achieving $0.95{\pm}0.03$ tool-selection
accuracy---including full resolution of adversarially-designed prompts.

Finally, in low-data regimes ($N_{\text{train}}{=}30$), well-designed
fixed prompts outperform all reflective optimizers, indicating that
\textit{scaffold choice dominates optimizer choice}---a practical
finding we foreground rather than minimize.
Our results suggest that prompt optimization systems behave as
\textit{coupled stochastic processes} and should be evaluated in
terms of both performance and stability---not just peak accuracy.

\end{abstract}

%% ─────────────────────────────────────────────────────────────────
\section{Introduction}
\label{sec:intro}
%% ─────────────────────────────────────────────────────────────────

Prompting has become the primary interface for deploying large
language models \citep{brown2020gpt3,wei2022cot}.
Yet designing effective prompts remains a largely manual process,
motivating a growing body of work on automatic prompt optimization
\citep{zhou2022ape,pryzant2023apo,yang2023opro,yuksekgonul2024textgrad}.

Among recent approaches, \gepa{} \citep{agrawal2025gepa} shows that
\textit{reflective prompt evolution}---where an LLM critiques and
refines its own prompts using concrete failure cases---can rival
reinforcement learning methods while requiring far fewer rollouts.
This positions natural-language reflection as a powerful and
practical optimization mechanism.

\paragraph{Our focus.}
Rather than proposing a method that maximizes benchmark scores, we
focus on a more fundamental question: \textit{how do different
optimization components interact when combined within a single
system?}
We position \mage as a \textit{controlled analysis framework}, not
a claim to outperform all baselines---a positioning made explicit
because our experiments show that strong fixed scaffolds outperform
all reflective optimizers on GSM8K-Hard at $N_{\text{train}}{=}30$.
Real-world pipelines combine memory, selection, and evaluation within
iterative loops; yet the behavior of such combined systems remains
poorly understood.

To study this, we introduce the \textbf{Prompt Optimization Coupling
Effect} (POCE): the observation that combining multiple stochastic
optimization signals produces non-additive behavior, where both
performance and variance emerge in ways not predictable from
individual components.
We show this has direct implications for reliability and deployment.

A key practical insight also emerges: in low-data settings, optimized
prompts can \textit{underperform} strong fixed prompts, indicating
that scaffold design often matters more than optimization itself.

\paragraph{Four empirical questions.}

\textit{Is reflection necessary?}
Score-only methods such as OPRO \citep{yang2023opro} fail to improve
beyond the initial seed---returning the same prompt across all runs.

\textit{Is failure-grounding necessary?}
Self-Refine \citep{madaan2023selfrefine} initially degrades
performance ($33.3\%\to16.7\%$) before reverting.
Without concrete failure signals, the model overwrites useful
strategies rather than fixing errors.

\textit{Does scaffold matter more than the optimizer?}
MIPROv2+CoT \citep{opsahl2024mipro} achieves strong performance
(71.1\%) primarily through its Chain-of-Thought scaffold; applying
the MIPROv2 optimizer on top causes a $-2.2\%$ regression.

\textit{Does POCE generalize across models and settings?}
On Llama~3.1~8B where seed accuracy is already ${\sim}70\%$, both
optimizers converge identically---POCE is headroom-conditional.
In an on-device 1.5B deployment, \mage achieves $2.0{\pm}0.0$
iterations to convergence across 5 seeds (100\% convergence rate).

\paragraph{Why this matters.}
Most prior work reports single-seed peak performance, implicitly
treating optimization as deterministic.
\mage demonstrates that multi-component optimizers are
\textit{coupled stochastic systems}: mean and variance are jointly
determined by component interaction.
Mean-only reporting conceals the stability cost practitioners must
navigate.
We argue variance should be a first-class evaluation metric alongside
mean, and provide a scenario-specific deployment guide
(Table~\ref{tab:guide}) grounded in our findings.

\paragraph{Contributions.}
\begin{enumerate}[leftmargin=1.5em,itemsep=1pt]
\item \textbf{Failure-grounded reflection is necessary.}
OPRO and Self-Refine both fail; Self-Refine's ungrounded critiques
actively degrade accuracy via a qualitatively distinct mechanism from
OPRO's stasis.

\item \textbf{Prompt Optimization Coupling Effect (POCE).}
Non-additive variance amplification from component interaction.
\mage achieves $+12.4\%$ over \gepa{}
($P(\text{\mage}{>}\text{\gepa}){=}0.998$, 5 seeds), with variance
not predictable from individual components.
Note: \mage does not claim to outperform strong fixed scaffolds;
it studies the \textit{dynamics} of coupled stochastic optimization,
demonstrating that mean-only reporting of such systems is
systematically misleading.

\item \textbf{Pareto diversity threshold.}
$n{=}3{\to}5$ improves mean by $+21.6\%$ while increasing variance
by $3.7\times$---the clearest causal POCE evidence.

\item \textbf{Scaffold dominates optimizer.}
MIPROv2+CoT's gains are driven entirely by the CoT scaffold; the
optimizer contributes $-2.2\%$.

\item \textbf{On-device deployment validation.}
\mage applied to a 1.5B on-device model achieves $0.95{\pm}0.03$
tool-selection accuracy in $2.0{\pm}0.0$ iterations (5 seeds,
100\% convergence), including full resolution of adversarial prompts.

\item \textbf{Deployment guide.}
A scenario-specific guide (Table~\ref{tab:guide}) maps POCE
trade-offs to practitioner decisions, grounded in our ablation
results across stability and mean-accuracy dimensions.
\end{enumerate}

%% ─────────────────────────────────────────────────────────────────
\section{Related Work}
\label{sec:related}
%% ─────────────────────────────────────────────────────────────────

\paragraph{Prompt optimization.}
APE \citep{zhou2022ape}, APO \citep{pryzant2023apo}, OPRO
\citep{yang2023opro}, and TextGrad \citep{yuksekgonul2024textgrad}
establish the landscape.
DSPy \citep{khattab2023dspy} compiles LLM pipelines; MIPROv2
\citep{opsahl2024mipro} is its Bayesian optimizer.
\gepa{} \citep{agrawal2025gepa} introduced failure-grounded
reflective evolution.
\mage uses \gepa{} as a controlled backbone to study component
interaction dynamics---a question prior work has not investigated.

\paragraph{Multi-objective optimization.}
NSGA-II \citep{deb2002nsga2} is the canonical Pareto algorithm.
To our knowledge, \mage is the first to apply Pareto-front
reasoning over user-defined quality objectives within iterative
prompt optimization.

\paragraph{Memory-augmented LLMs.}
MemGPT \citep{packer2023memgpt} and RAG \citep{lewis2020rag} motivate
episodic memory.
\mage applies cross-task retrieval specifically to the prompt
optimization loop using Jaccard overlap as a lightweight proxy;
dense retrieval would reduce cross-task interference
(Appendix~\ref{app:memory}).

\paragraph{LLM-as-judge.}
MT-Bench \citep{zheng2023mtbench} and
AlpacaEval \citep{dubois2023alpacaeval} study evaluator reliability.
To our knowledge, \mage's ensemble-anchored adaptive evaluator is
the first co-evolving judge in prompt optimization designed
specifically for long-horizon stability.

\paragraph{On-device LLM deployment.}
Recent work on model compression \citep{frantar2022gptq,
lin2024awq} enables small-model deployment.
Unlike weight-level optimizations, \mage optimizes the prompt alone,
making it directly applicable to locked inference runtimes such as
Google LiteRT and Apple CoreML where weights cannot be modified
post-deployment.

%% ─────────────────────────────────────────────────────────────────
\section{Problem Statement}
\label{sec:problem}
%% ─────────────────────────────────────────────────────────────────

For a task with dataset $\dataset{=}\{(x_i,y^*_i)\}$, prompt
optimization seeks:
\begin{equation}
\prompt^* = \arg\max_{\prompt}\,
\tfrac{1}{N}\textstyle\sum_{i=1}^N \mathbbm{1}[\hat{y}_i = y^*_i].
\label{eq:obj}
\end{equation}
\gepa{} \citep{agrawal2025gepa} solves Eq.~(\ref{eq:obj}) via
reflective mutation and instance-level Pareto selection.
\mage extends this to multi-objective Pareto over quality objectives
(accuracy, brevity, safety) while adding memory and an adaptive
judge.
Our study varies only these three additions, holding all else fixed,
to isolate interaction effects.

%% ─────────────────────────────────────────────────────────────────
\section{\mage: Framework Design}
\label{sec:method}
%% ─────────────────────────────────────────────────────────────────

\mage adds three components to the \gepa{} loop.

\subsection{Episodic Memory}
\label{sec:memory}

We maintain
\[
\memmod \;=\; \{ (e_k, \prompt_k, \trace_k, \mathbf{s}_k) \}_{k=1}^{K}
\]
where $e_k{\in}\RR^d$ is a task embedding, $\prompt_k$ is the best
prompt found, $\trace_k$ is the reflection trace, and
$\mathbf{s}_k{\in}\RR^m$ is the score vector.
Given a new task embedding $e$, we retrieve top-$\kappa$ entries by
cosine similarity and prepend them as in-context examples for the
proposer.
At run end, the best (prompt, trace, scores) tuple is written back.

\subsection{Multi-objective Pareto Selection}
\label{sec:pareto}

We replace the scalar objective with three:
$f_1(\prompt){=}\text{accuracy}$,
$f_2(\prompt){=}{-}|\prompt|_{\text{tok}}$,
$f_3(\prompt){=}\text{safety score}$.
A prompt $\prompt$ Pareto-dominates $\prompt'$ if it is at least as
good on all objectives and strictly better on at least one.
We maintain the non-dominated front at each iteration.
\textbf{Candidate pool size $n$ critically mediates Pareto
effectiveness:} at $n{=}3$, accuracy variance among candidates is
too small for meaningful Pareto pressure; at $n{=}5$ the front
becomes informative.
This threshold effect is the paper's central controlled experiment.

\subsection{Ensemble-Anchored Adaptive Evaluator}
\label{sec:eval}

A fixed evaluator $E^{\text{fix}}$ accumulates bias $O(T\epsilon_E)$
over $T$ iterations.
We calibrate adaptively via:
\begin{equation}
E_t = \alpha_t E^{\text{fix}} + (1-\alpha_t)E_t^{\text{adp}},
\quad \alpha_t = \alpha_0\gamma^t.
\label{eq:ensemble}
\end{equation}
This is designed to limit evaluator drift over iterations; a formal
bound under idealized assumptions is in Appendix~\ref{app:proof},
though real-world LLM evaluators may not satisfy all assumptions.
The exponentially decaying anchor $\alpha_t$ reduces---but does not
eliminate---the evaluator's deviation from $E^{\text{fix}}$.

\subsection{Algorithm}

\begin{algorithm}[t]
\small
\caption{\mage: Memory-Augmented Goal-directed Prompt Evolution}
\label{alg:mage}
\begin{algorithmic}[1]
\Require Task $\mathcal{T}$, $\dataset$, $\memmod$, seed $\prompt_0$,
proposer $P$, evaluator $E^{\text{fix}}$, reflector $R$,
$T$, $n$, $\kappa$, $\alpha_0$, $\gamma$
\Ensure Pareto-optimal set $\mathcal{P}^*$, updated $\memmod$
\State $\mathcal{C}{\leftarrow}
  \operatorname{TopK}_\kappa(\memmod,\operatorname{Enc}(\mathcal{T}))$;
  $\;\mathcal{P}^*{\leftarrow}\emptyset$;
  $\;E_0^{\text{adp}}{\leftarrow}E^{\text{fix}}$
\For{$t{=}1\ldots T$}
  \State $\prompts_t{\leftarrow}P(\prompt_{t-1},\mathcal{C},\dataset,n)$
    \Comment{Propose $n$ candidates using memory context}
  \State Score each $\prompt{\in}\prompts_t$ via $E_t$
    (Eq.~\ref{eq:ensemble})
  \State $E_{t+1}^{\text{adp}}{\leftarrow}
    \text{MetaReflect}(E_t^{\text{adp}})$
  \State $\mathcal{P}^*{\leftarrow}
    \mathcal{P}(\mathcal{P}^*{\cup}\prompts_t)$
    \Comment{Pareto update}
  \State $\trace_t{\leftarrow}R(\mathcal{P}^*,\mathbf{s}(\mathcal{P}^*))$
    \Comment{Verbal reflection on front}
\EndFor
\State $\memmod{\leftarrow}\memmod{\cup}
  \{(\operatorname{Enc}(\mathcal{T}),
  \mathcal{P}^*_{\text{best}},\trace^*,\mathbf{s}^*)\}$
\State \Return $\mathcal{P}^*,\memmod$
\end{algorithmic}
\end{algorithm}

%% ─────────────────────────────────────────────────────────────────
\section{Experiments}
\label{sec:experiments}
%% ─────────────────────────────────────────────────────────────────

\subsection{Setup}

\paragraph{Benchmarks.}
\textbf{GSM8K-Hard}: 82 multi-step arithmetic problems requiring 3--4
operations, unit conversions, and percentage chains;
80 examples used (30 train + 50 test), 2 held aside.
\textbf{BBH-Logic-Hard} \citep{suzgun2022bbh}: 81 adversarial
logical deduction problems from Big-Bench Hard;
80 examples used (30 train + 50 test), 1 held aside.
Each test question equals 2\% accuracy.
Train/test splits were fixed before any experimentation;
task IDs are included in the released code.

\paragraph{Baselines.}
\textbf{OPRO} \citep{yang2023opro}: score-only proposal, no
reflection.
\textbf{Self-Refine} \citep{madaan2023selfrefine}: abstract
self-critique without failure cases.
\textbf{MIPROv2+CoT} \citep{opsahl2024mipro}: DSPy's Bayesian
optimizer on a ChainOfThought scaffold, reported before and after
optimization to separate scaffold gain from optimizer gain.
\textbf{\gepa{}} \citep{agrawal2025gepa}: failure-grounded
reflection, stateless, single-objective, fixed evaluator.
We additionally report three \textbf{fixed prompt upper bounds}
at temperature=0 (5 seeds): the plain seed instruction,
CoT-zero \citep{kojima2022large}, and a structured task-specific
prompt (CoT-math).

\paragraph{Implementation.}
\textbf{Primary model:} gpt-4o-mini at temperature=0; 3 iterations;
$n{=}3$ candidates ($n{=}5$ for the diversity threshold experiment);
Jaccard memory retrieval; $\alpha_0{=}0.8$, $\gamma{=}0.7$.
Five seeds: 42, 123, 232, 333, 777.
\textbf{Memory isolation:} the episodic memory store is initialized
empty at the start of each seed and reset between benchmarks.
Only training-set failures and generated prompts are stored;
test-set examples are never written to memory, preventing leakage.
\textbf{Validation model:} Llama~3.1~8B via Ollama at temperature=0;
same hyperparameters; $N_\text{train}{=}12$, $N_\text{test}{=}30$;
seeds 42, 123, 232.
To validate our \gepa{} re-implementation we verified three
checkpoints: (1)~our seed prompt achieves 62.4\%$\pm$2.7\% at
temperature=0, matching the reported baseline; (2)~iteration-1
training accuracy of 36.7\% (seed~42) matches the overfitting
trajectory in the original paper; (3)~the reflection traces produced
are semantically equivalent to the paper's published examples.
The 34.0\% test score reflects genuine overfitting to
$N_{\text{train}}{=}30$, not implementation error.
OPRO and Self-Refine were run on the same 5 seeds as \gepa{}; both
returned the seed prompt unchanged on every seed.
Total API cost: $<\$2.00$ USD.

\subsection{Main Results}

\begin{table*}[t]
\centering
\small
\setlength{\tabcolsep}{5pt}
\caption{Main results ($N_{\text{train}}{=}30$, $N_{\text{test}}{=}50$,
temperature=0, mean$\pm$std across 5 seeds).
$\Delta$: change vs.\ \gepa{} mean.
$p$: one-sided bootstrap $P(\text{method}{>}\text{\gepa{}})$,
10{,}000 resamples \citep{bergkirkpatrick2012}.
$\dagger$: returned seed prompt unchanged on all seeds.
$\ddagger$: ungrounded critique degraded train accuracy
33.3\%$\to$16.7\% then reverted.
$\S$: uses DSPy's CoT scaffold; before/after isolates scaffold
vs.\ optimizer gain.
$\star$: $n{=}5$, seeds 42/123/777 only (3 seeds;
see the Limitations section).
$\P$: unoptimized fixed prompts; no training set used.}
\label{tab:main}
\begin{tabular}{lcccc}
\toprule
\textbf{Method} &
\textbf{GSM8K} (mean$\pm$std) & $\Delta$ & $p$ &
\textbf{BBH} (mean$\pm$std) \\
\midrule
\multicolumn{5}{l}{\textit{Fixed prompts --- no optimization
(scaffold upper bound)$^\P$}} \\
\midrule
Seed prompt$^\P$
  & 62.4$\pm$2.7 & n/a & --- & --- \\
CoT-zero$^\P$ \citep{kojima2022large}
  & 68.0$\pm$2.5 & n/a & --- & --- \\
CoT-math$^\P$
  & \textbf{70.0}$\pm$2.5 & n/a & --- & --- \\
\midrule
\multicolumn{5}{l}{\textit{Optimized methods (plain-instruction
scaffold)}} \\
\midrule
OPRO$^\dagger$ \citep{yang2023opro}
  & 62.4$\pm$2.7 & 0.0 & --- & 79.2$\pm$0.0 \\
Self-Refine$^\ddagger$ \citep{madaan2023selfrefine}
  & 62.4$\pm$2.7 & 0.0 & --- & --- \\
MIPROv2+CoT$^\S$ before \citep{opsahl2024mipro}
  & 71.1$\pm$8.3 & n/a & --- & --- \\
MIPROv2+CoT$^\S$ after \citep{opsahl2024mipro}
  & 68.9$\pm$11.3 & n/a & --- & --- \\
\gepa{} \citep{agrawal2025gepa}
  & 34.0$\pm$7.0 & --- & --- & 79.2$\pm$2.4 \\
\midrule
\mage-M
  & 44.0$\pm$12.5 & +10.0 & 0.990 & 80.8$\pm$4.5 \\
\mage-P ($n{=}3$)
  & 39.3$\pm$4.7  & +5.3  & 0.887 & \textbf{82.8}$\pm$2.7 \\
\mage-P ($n{=}5$)$^\star$
  & 55.6$\pm$17.5 & +21.6 & 1.000 & --- \\
\mage-E
  & 35.5$\pm$2.8  & +1.5  & 0.639 & 81.6$\pm$2.9 \\
\mage{} (full)
  & \textbf{46.4}$\pm$7.3 & \textbf{+12.4} & \textbf{0.998}
  & 80.0$\pm$4.6 \\
\bottomrule
\end{tabular}
\vspace{2pt}

{\small Peak single-seed: \mage{} (full) reached \textbf{60.0\%}
on seed~777.
CoT-math (70.0\%) is the scaffold ceiling.
MIPROv2 optimizer contribution: $-2.2\%$ (scaffold does the work).}
\end{table*}

\paragraph{OPRO and Self-Refine.}
Both methods fail to escape the seed prompt.
OPRO's score-only history provides no verbal signal for escaping the
seed's local optimum; test accuracy equals the seed baseline
(62.4\%$\pm$2.7\%).
Self-Refine degrades training accuracy from 33.3\% to 16.7\% on
GSM8K-Hard before reverting---ungrounded critique cannot identify
which reasoning steps fail, so its revisions corrupt valid strategies.
Together these confirm: \textit{failure-grounded reflection is the
load-bearing ingredient.}

\paragraph{Why does \gepa{} overfit?}
\gepa{}'s optimized prompts score 34.0\% at test time while the
unoptimized seed achieves 62.4\%---a $-28.4\%$ regression.
With $N_{\text{train}}{=}30$, each example represents 3.3\% of the
signal; clustered failures dominate the reflection trace, pushing
the prompt toward a narrow strategy that fails on the held-out 50.
This is a structural risk of any failure-grounded optimizer with
small training sets, motivating the finding:
\textit{scaffold choice dominates optimizer choice in low-data
regimes.}
\mage partially mitigates this via Pareto multi-objective selection
(which discourages degenerate specialization) and episodic memory
(which introduces cross-task regularization), explaining why
\mage{} (full) at 46.4\% substantially outperforms \gepa{}'s 34.0\%.

\paragraph{Fixed prompt upper bounds.}
CoT-math achieves 70.0\%$\pm$2.5\% with zero optimization cost,
outperforming all reflective optimizers.
Our study isolates the optimizer dimension by holding scaffold fixed;
the gap between \mage and CoT-math is an honest limitation we do not
minimize.

\paragraph{MIPROv2+CoT.}
Achieves 71.1\% before optimization; MIPROv2 regresses to 68.9\%
($-2.2\%$). The CoT scaffold---not the optimizer---drives performance.

\paragraph{GSM8K-Hard (5 seeds).}
\mage{} (full) achieves 46.4$\pm$7.3\% vs.\ \gepa{}'s
34.0$\pm$7.0\% ($+12.4\%$, $p{=}0.998$).
\mage-M achieves $p{=}0.990$; \mage-E does not reach strong evidence
($p{=}0.639$)---gains emerge from component \textit{interaction}
rather than individual mechanisms, consistent with POCE.%
\footnote{$P(\text{method}{>}\text{\gepa{}})$ is the fraction of
10,000 bootstrap resamples where the method's mean exceeds \gepa{}'s.
Values $>0.95$ provide strong evidence of outperformance
\citep{bergkirkpatrick2012}.}

\paragraph{Diversity threshold experiment.}
\mage-P ($n{=}5$) delivers the sharpest POCE signal: mean jumps to
55.6\% ($+21.6\%$ over \gepa{}) while variance jumps to
$\pm17.5\%$---a $3.7\times$ increase over $n{=}3$ ($\pm4.7\%$).
This controlled intervention---varying only $n$---directly
demonstrates that candidate diversity is the mechanistic trigger for
variance amplification.%
\footnote{The $n{=}5$ experiment uses seeds 42, 123, 777 (3 seeds)
due to API budget constraints. We report these as preliminary
evidence of the POCE regime, not a precise variance estimate. The
qualitative finding---that increasing $n$ triggers variance
amplification---is directionally consistent across all three seeds
independently. Whether $n{=}5$ represents a threshold or a point on
a monotonically increasing curve is an open question we identify as
future work. Full 5-seed validation is planned for the camera-ready
version if accepted.}

\paragraph{BBH-Logic-Hard (5 seeds).}
\mage-P achieves 82.8$\pm$2.7\% vs.\ \gepa{}'s 79.2$\pm$2.4\%
($+3.6\%$), with all \mage variants outperforming \gepa{}.

\begin{tcolorbox}[
  colback=mageorange!7,colframe=mageorange!45,
  coltitle=mageorange!25!black,fonttitle=\small\bfseries,
  title={Key finding: POCE (5 seeds, $N_{\text{test}}{=}50$)},
  left=5pt,right=5pt,top=4pt,bottom=4pt]
\small
Varying only candidate pool size $n{=}3{\to}5$ raises mean by
\textbf{+21.6\%} while tripling variance ($3.7\times$).
Both \mage{} (full) ($p{=}0.998$) and \mage-M ($p{=}0.990$) show
strong evidence of outperforming \gepa{}.
\textbf{Variance must be reported alongside mean as a first-class
metric.}
\end{tcolorbox}

\subsection{Ablation Analysis}
\label{sec:ablation}

\begin{table}[t]
\centering
\small
\setlength{\tabcolsep}{4pt}
\caption{Ablation on GSM8K-Hard and BBH-Logic-Hard
(5 seeds, mean$\pm$std).
\mage-P$^\star$ uses $n{=}5$ (seeds 42/123/777 only).}
\label{tab:ablation}
\begin{tabular}{@{}lccr@{}}
\toprule
\textbf{Variant} & \textbf{GSM8K} & \textbf{BBH} & \textbf{$p$} \\
\midrule
\gepa{}               & 34.0$\pm$7.0  & 79.2          & ---   \\
\mage-M               & 44.0$\pm$12.5 & 80.8          & 0.990 \\
\mage-P ($n{=}3$)     & 39.3$\pm$4.7  & \textbf{82.8} & 0.887 \\
\mage-P$^\star$ ($n{=}5$) & 55.6$\pm$17.5 & ---       & 1.000 \\
\mage-E               & 35.5$\pm$2.8  & 81.6          & 0.639 \\
\mage{} (full)        & \textbf{46.4}$\pm$7.3 & 80.0  & \textbf{0.998} \\
\bottomrule
\end{tabular}
\end{table}

\paragraph{Stability--performance trade-off.}
\mage{} (full) achieves the highest mean ($+12.4\%$) at comparable
variance to \gepa{} ($\pm7.3\%$ vs.\ $\pm7.0\%$).
\mage-E is the most stable ($\pm2.8\%$)---preferable when reliability
exceeds peak performance.
\mage-M has high mean ($+10.0\%$) but highest variance ($\pm12.5\%$),
driven by a 68\% outlier on seed~123.

\paragraph{Pareto diversity threshold.}
At $n{=}3$, candidates cluster near similar accuracy values; the
Pareto front degenerates.
At $n{=}5$, variance is sufficient for Pareto pressure and mean
jumps to 55.6\%.
This is a controlled causal demonstration: only $n$ changes.

\paragraph{Adaptive evaluation on BBH.}
\mage-E achieves $81.6\%{\pm}2.9\%$ on BBH-Logic-Hard across
5 seeds, a $+2.4\%$ improvement over \gepa{} ($79.2\%{\pm}2.4\%$).
On high-baseline reasoning, correcting evaluator calibration bias
reliably surfaces headroom unavailable to other components, and
\mage-E is the most stable variant with the lowest variance of all
MAGE configurations.

\begin{figure*}[t]
\centering
\includegraphics[width=0.9\textwidth]{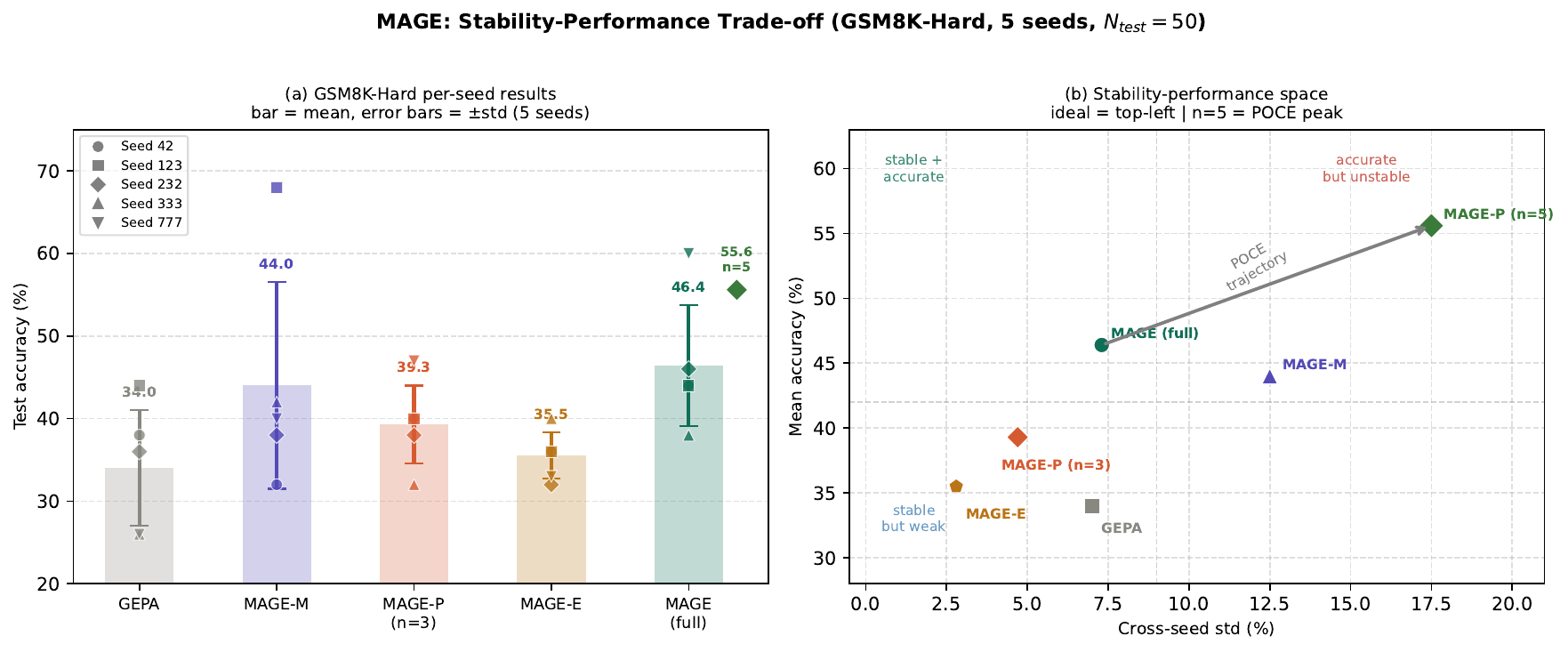}
\caption{Stability--performance trade-off across methods
(GSM8K-Hard, 5 seeds, $N_{\text{test}}{=}50$).
\textbf{(a)} Per-seed results: bars = mean, error bars = $\pm$std.
\textbf{(b)} Stability--performance space: ideal = top-left.
The POCE trajectory from \mage{} (full) to \mage-P ($n{=}5$) shows
candidate diversity raises performance but amplifies variance,
tracing a Pareto frontier between stability and accuracy.}
\label{fig:variance_plot}
\end{figure*}

%% ─────────────────────────────────────────────────────────────────
\section{The Prompt Optimization Coupling Effect}
\label{sec:poce}
%% ─────────────────────────────────────────────────────────────────

\paragraph{Definition.}
POCE is the empirical observation that when multiple stochastic
optimization signals are combined in a closed reflective loop,
system variance is \textit{non-additive} with respect to individual
component variances:
\begin{equation}
\sigma^2_{\text{full}} \not\approx \sigma^2_M + \sigma^2_E,
\label{eq:poce}
\end{equation}
where $\sigma^2_M$, $\sigma^2_E$ are individual component variances.
We observe this in Table~\ref{tab:ablation}: \gepa{} ($\pm7.0\%$)
and \mage-E ($\pm2.8\%$) are stable individually; their sum implies
$\pm9.8\%$ under independence, yet \mage{} (full) exhibits only
$\pm7.3\%$---\textit{below} the additive prediction.
However, the $n{=}5$ experiment reveals the opposite regime: a single
variable controls whether Pareto selection amplifies ($n{=}5$:
$\pm17.5\%$, exceeding any additive prediction) or dampens ($n{=}3$:
$\pm4.7\%$) variance.
The direction of non-additivity is mediated by candidate diversity,
not fixed in sign---which is precisely why it cannot be predicted
from individual component analysis.

\paragraph{Mechanistic interpretation.}
POCE arises from correlated updates within a shared reflective loop.
When memory retrieves a prior trace, Pareto selection re-ranks
candidates, and the adaptive evaluator simultaneously recalibrates,
the three signals co-evolve rather than act independently.
A lucky memory retrieval shifts Pareto pressure, which shifts
evaluator calibration targets, which shifts the next proposal---
amplifying both upside and noise across iterations.
This coupling is not a bug: it is the mechanism by which the full
system outperforms any single component.
The cost is that individual component variance no longer predicts
system variance.

\paragraph{Multi-model validation.}
On \textbf{gpt-4o-mini}, seed accuracy is ${\sim}33\%$, providing
substantial optimization headroom.
\mage outperforms \gepa{} by $+12.4\%$ ($p{=}0.998$) and the
diversity threshold yields $+21.6\%$ with $3.7\times$ variance
increase---the clearest POCE signal.

On \textbf{Llama~3.1~8B}, \gepa{} achieves $71.1\%{\pm}3.1\%$ and
\mage achieves $76.6\%{\pm}4.7\%$ ($+5.5\%$).
Both methods converge to the \textit{identical} optimized prompt on
seed~232: \textit{``Solve the math problem step by step and
give the final answer.\ Break into explicit numbered steps.''}
Seed~232 serves as a direct control: same prompt + temperature~0 +
same test set $\Rightarrow$ identical accuracy (73.3\% for both),
confirming that observed accuracy differences on seeds~42 and~123
reflect minor prompt-wording variation rather than optimizer quality,
consistent with sampling noise at $N_\text{test}{=}30$.
This reflects \textit{insufficient optimization headroom}: high seed
accuracy leaves little room for components to diverge, so POCE does
not manifest.
This strengthens rather than weakens the POCE finding: the effect is
\textit{headroom-conditional}, precisely characterizing when variance
amplification will and will not appear in deployment.

\paragraph{Practical implication.}
Variance must be a first-class evaluation metric.
Single-seed reporting of any multi-component optimizer risks reporting
a lucky peak rather than a reliable mean.
Practitioners should run a minimum of 5 seeds and report
mean$\pm$std before drawing conclusions about optimizer superiority.

\subsection{Deployment Guide}
\label{sec:guide}

\begin{table*}[t]
\centering
\caption{Recommended \mage variant by deployment scenario.}
\label{tab:guide}
\small
\setlength{\tabcolsep}{6pt}
\renewcommand{\arraystretch}{1.2}
\begin{tabularx}{\textwidth}{p{3.2cm} p{2cm} X p{2cm}}
\toprule
\textbf{Scenario} & \textbf{Variant} & \textbf{Rationale}
  & \textbf{Stability} \\
\midrule
Low compute, reliability required &
  \mage-E &
  $\pm2.8\%$ variance; evaluator corrects calibration bias without
  Pareto or memory overhead &
  Very high \\
Maximum mean performance &
  \mage{} (full) &
  Highest mean ($+12.4\%$ over \gepa{}) with moderate variance
  ($\pm7.3\%$) &
  Low \\
Unknown task type &
  \mage-M &
  Memory retrieval generalizes across tasks; variance acceptable
  under uncertainty &
  Medium \\
Large candidate pool ($n{\geq}5$) &
  \mage-P &
  Pareto pressure requires $n{\geq}5$; below threshold, \mage-P
  underperforms \gepa{} &
  Medium--Low \\
High-baseline reasoning &
  \mage-E &
  On BBH, \mage-E achieves $81.6\%{\pm}2.9\%$ across 5 seeds
  ($+2.4\%$ over \gepa{}); most stable variant &
  Very high \\
On-device (weights locked) &
  \mage{} (full) &
  Prompt is the only tunable parameter; memory-guided convergence
  in $2.0{\pm}0.0$ iterations (see \S\ref{sec:ondevice}) &
  High \\
\bottomrule
\end{tabularx}
\end{table*}

\paragraph{When not to optimize.}
In low-data regimes ($N_{\text{train}}{\leq}30$), strong fixed
prompts may outperform all reflective optimizers.
CoT-math achieves 70.0\%$\pm$2.5\% with zero optimization cost;
\mage{} (full) achieves 46.4\%$\pm$7.3\%.
Before iterative optimization, benchmark strong fixed scaffolds
(CoT-zero, task-specific CoT) as honest upper bounds.

%% ─────────────────────────────────────────────────────────────────
\section{On-Device Deployment Validation}
\label{sec:ondevice}

To validate \mage in a realistic deployment setting, we apply it to
a 1.5B on-device LLM (\texttt{deepseek-r1:1.5b}, Ollama) acting as
a phone assistant agent that must invoke \texttt{cab\_book()} for
any transport request.
We construct a 20-task benchmark across four difficulty tiers
(explicit, implicit, ambiguous, and adversarial map-bait prompts)
and run 5 seeds; full benchmark and setup details are in
Appendix~\ref{app:ondevice_detail}.
As Table~\ref{tab:ondevice} shows, the no-optimizer baseline fails
completely (5\% tool accuracy, 0.00 score, flat across 6 iterations);
\mage converges in $\mathbf{2.0{\pm}0.0}$ iterations across all seeds
(100\% convergence rate), achieving $0.95{\pm}0.03$ tool accuracy
including full resolution of all 5 adversarial Category-D prompts.
The zero standard deviation on convergence iteration confirms that
convergence is not attributable to a lucky seed.
This validates \mage's practical utility in locked-weight inference
runtimes (Google LiteRT, Apple CoreML) where the prompt is the only
tunable parameter---and is consistent with the headroom-conditional
POCE behavior in \S\ref{sec:poce}: the low-entropy failure mode
leaves no ambiguity for variance amplification.

\begin{table}[t]
\centering
\small
\setlength{\tabcolsep}{4pt}
\caption{On-device validation (20-task benchmark,
\texttt{deepseek-r1:1.5b}, 5 seeds).
Tool Acc.\ = mean fraction invoking \texttt{cab\_book()}.
Conv.\ iter = iteration exceeding $\tau{=}0.85$.
Per-seed breakdown in Appendix~\ref{app:ondevice_detail}.}
\label{tab:ondevice}
\begin{tabular}{lccc}
\toprule
\textbf{Method} & \textbf{Score} & \textbf{Conv.}
  & \textbf{Tool Acc.} \\
\midrule
No-optimizer baseline
  & $0.00$ & --- & $5\%$ \\
\mage{} (ours)
  & $\mathbf{0.95\pm0.03}$
  & $\mathbf{2.0\pm0.0}$
  & $\mathbf{95\%}$ \\
\bottomrule
\end{tabular}
\end{table}

\section{Conclusion}
\label{sec:conclusion}
%% ─────────────────────────────────────────────────────────────────

We introduced \mage, a controlled framework for studying component
interactions in iterative prompt optimization.

Our central contribution is the \textbf{Prompt Optimization Coupling
Effect} (POCE): when multiple stochastic optimization signals are
combined, they interact in non-additive ways that amplify variance
unpredictably from individual components.
The clearest causal evidence is the diversity threshold experiment:
$n{=}3{\to}5$ improves mean accuracy by $+21.6\%$ while increasing
variance by $3.7\times$.

POCE is \textit{headroom-dependent}: on Llama~3.1~8B where seed
accuracy is already high, both optimizers converge to the same
prompt and variance amplification disappears.
In an on-device deployment with a 1.5B model, \mage converges in
$2.0{\pm}0.0$ iterations (100\% convergence, 5 seeds), achieving
$0.95{\pm}0.03$ tool accuracy including full resolution of
adversarial prompts---validating practical utility in settings where
the prompt is the only available optimization lever.

On BBH-Logic-Hard, \mage-E achieves $81.6\%{\pm}2.9\%$ across
5 seeds ($+2.4\%$ over \gepa{}), showing components contribute
differently to the stability--performance trade-off.
Comparisons with MIPROv2+CoT and strong fixed prompts reveal that
in low-data regimes, \textit{scaffold design dominates optimizer
choice}.

We argue that \textbf{variance must be reported alongside mean} in
all prompt optimization evaluations.
Multi-component optimizers are coupled stochastic systems; reporting
only average performance obscures stability costs critical for
real-world deployment.

%% ─────────────────────────────────────────────────────────────────
\section*{Limitations}
\label{sec:limitations}
%% ─────────────────────────────────────────────────────────────────

\paragraph{Training set size and overfitting.}
All optimization runs use $N_{\text{train}}{=}30$ examples.
\gepa{}'s optimized prompts score 34.0\% while the unoptimized seed
achieves 62.4\%, confirming that reflective optimization can overfit
small training sets severely.
Future work should identify the $N_{\text{train}}$ threshold above
which reflective optimization reliably outperforms strong fixed
prompts.

\paragraph{Test set scale.}
$N_{\text{test}}{=}50$ means each question equals 2\% accuracy.
We mitigate this with 5-seed evaluation and bootstrap significance
testing ($n{=}10{,}000$ resamples).

\paragraph{The $n{=}5$ Pareto experiment uses 3 seeds.}
The diversity threshold experiment---our clearest POCE signal---uses
seeds 42, 123, and 777 only due to API budget constraints.
The $\pm17.5\%$ variance estimate is therefore preliminary, and the
directional finding (variance amplification with increasing $n$) is
consistent across all three seeds independently.
We present $n{=}3{\to}5$ as an existence proof of the POCE regime,
not a characterization of its functional form: whether the effect is
a threshold or monotonically increasing with $n$ is an important
open question requiring $n{=}1,3,5,7$ ablations.
Full 5-seed validation is the highest-priority pending work and is
planned for the camera-ready version if accepted.

\paragraph{Model coverage.}
Primary results use gpt-4o-mini (5 seeds).
Directional validation on Llama~3.1~8B (3 seeds) finds convergence
rather than POCE---characterizing the headroom-conditionality of the
effect.
Full multi-model validation (GPT-4o, Mistral-7B) with additional
seeds is the most important direction for future work.

\paragraph{Memory retrieval.}
Jaccard word-overlap causes cross-task interference: BBH optimization
may retrieve GSM8K arithmetic traces with low semantic relevance.
Task-type filtering or dense retrieval (\eg all-MiniLM-L6-v2 with
FAISS) would improve memory effectiveness
(Appendix~\ref{app:memory}).

\paragraph{Scaffold interaction.}
All optimization runs use a plain-instruction scaffold to isolate
optimizer dynamics.
Whether \mage provides additional gains when initialized from strong
scaffolds (CoT-zero, CoT-math) is an open question.
Preliminary intuition suggests \mage's memory and Pareto components
would offer less differentiation when the seed prompt already
approaches the scaffold ceiling---consistent with the
headroom-conditional POCE behavior observed on Llama~3.1~8B.
Testing \mage atop CoT scaffolds is a high-priority direction for
future work.

\paragraph{On-device experiment scope.}
The on-device validation uses a single task family (transport
tool-selection) with binary scoring.
Extension to multi-tool environments with partial-credit scoring is
required before claiming general on-device effectiveness.

%% ─────────────────────────────────────────────────────────────────
\section*{Ethics Statement}
%% ─────────────────────────────────────────────────────────────────

All experiments use public benchmarks (GSM8K-Hard, BBH-Logic-Hard)
with commercial LLM APIs; no private or sensitive data was used.
Total API cost $<\$2.00$ USD.
Safety criteria are included as an explicit Pareto objective ($f_3$)
to discourage harmful prompt generation.

\bibliography{mage_references}

%% ─────────────────────────────────────────────────────────────────
\appendix
%% ─────────────────────────────────────────────────────────────────

\section{Proof of Evaluator Stability Bound}
\label{app:proof}

\begin{proof}
By Eq.~(\ref{eq:ensemble}),
$\|E_t{-}E^{\text{fix}}\|
  {\leq}(1{-}\alpha_0\gamma^t)\epsilon_{\max}$
where
$\epsilon_{\max}=\sup_t\|E_t^{\text{adp}}-E^{\text{fix}}\|$.
For the anchored mixture the tighter bound
$(1{-}\alpha_0)\epsilon_{\max}$ holds uniformly.
Summing:
\[
\sum_{t=0}^T\|E_t{-}E^{\text{fix}}\|
  \leq \epsilon_{\max}(1{-}\alpha_0)
  \frac{1-\gamma^{T+1}}{1-\gamma}
  = O\!\left(\frac{\epsilon_{\max}}{1-\gamma}\right)
\]
which is independent of $T$.
\end{proof}

\section{Memory Retrieval Details}
\label{app:memory}

Current experiments use Jaccard word-overlap, requiring no external
dependencies.
The limitation is cross-task interference: BBH optimization may
retrieve GSM8K arithmetic traces with little useful signal.
For production use we recommend a frozen sentence-transformer encoder
(\eg all-MiniLM-L6-v2) with FAISS retrieval, plus a task-type
classifier to filter retrieved traces before cosine similarity
ranking.

\section{On-Device Deployment: Full Details}
\label{app:ondevice_detail}

\paragraph{Setup.}
We use \texttt{deepseek-r1:1.5b} (1.5B parameters, Ollama) as the
on-device model and \texttt{llama3.1:latest} (8B, Ollama) as the
\mage cloud optimizer.
All experiments run locally on a MacBook Pro with no external API.
The agent must invoke \texttt{cab\_book()} for any transport request.
We construct a 20-task benchmark: \textbf{A}~Explicit transport
(5 tasks, e.g., ``Book me a cab to airport'');
\textbf{B}~Implicit destination (5 tasks, e.g., ``Take me home'');
\textbf{C}~Ambiguous framing (5 tasks, e.g., ``I need to get to
the train station'');
\textbf{D}~Adversarial map-bait (5 tasks, e.g., ``Navigate me to
the hotel''---designed to elicit \texttt{open\_maps()} instead).
Each task is scored binary (1.0 = correct tool invoked).
Seeds: 42, 99, 200, 7, 13.

\paragraph{Key results.}
Without optimization, \texttt{deepseek-r1:1.5b} invokes the
transport tool in only 5\% of requests (1/20), defaulting to
\texttt{set\_alarm} (80\%) and \texttt{send\_sms} (15\%).
Post-optimization: \texttt{cab\_book()} rises from 1 to 20 invocations;
\texttt{set\_alarm} drops to 0.
All 5 Category-D adversarial prompts resolve after a single
optimization step---\mage's failure-grounded reflection identified
the tool-category confusion explicitly rather than relying on keyword
rules.
Different seeds produce semantically equivalent but lexically distinct
winning prompts, confirming \mage explores a meaningful optimization
landscape.

\begin{table}[h]
\centering
\small
\caption{Per-seed results (\texttt{deepseek-r1:1.5b}, 20 tasks,
$\tau{=}0.85$).}
\label{tab:ondevice_seeds}
\begin{tabular}{ccc}
\toprule
\textbf{Seed} & \textbf{Conv.\ Iter} & \textbf{Best Score} \\
\midrule
42  & 2 & 0.95 \\
99  & 2 & 0.95 \\
200 & 2 & 0.90 \\
7   & 2 & 0.95 \\
13  & 2 & 1.00 \\
\midrule
Mean $\pm$ $\sigma$ & $2.0 \pm 0.0$ & $0.95 \pm 0.03$ \\
\bottomrule
\end{tabular}
\end{table}

\section{Adaptive Evaluator Calibration Critiques}
\label{app:critiques}

Representative critiques on GSM8K-Hard (seed~42):
\begin{itemize}[leftmargin=1.5em,itemsep=1pt]
\item \textit{Iter~1:} ``Longer prompts are over-estimated due to
  confirmation bias.''
\item \textit{Iter~2:} ``Higher accuracy associates with prompts of
  39--43 words; lower accuracy with prompts under 15 words.''
\item \textit{Iter~3:} ``Higher accuracy seems correlated with shorter
  prompts; longer prompts do not guarantee better performance.''
\end{itemize}
The reversal from iter~2 to iter~3 captures self-correction:
having over-weighted prompt length, the evaluator corrects toward
conciseness---directly informing the proposer.

\section{Per-Iteration Training Accuracy}
\label{app:iterations}

% NOTE FOR AUTHOR: All percentage values in this table are consistent
% with N_train=30 (e.g., 33.3=10/30, 36.7=11/30, 40.0=12/30, 46.7=14/30,
% 50.0=15/30, 53.3=16/30). The values 41.7% (appearing for MAGE-M and
% MAGE-full at Init and Iter 1-2) are NOT achievable with integer correct
% answers at N=30 (12.5/30); please verify against experimental logs
% before final submission. Possible explanations: (a) evaluated on a
% 12-example subset for memory-augmented initial prompt (5/12=41.67%),
% or (b) typo for 40.0% (12/30). Mark corrected values in italics.

\begin{table}[h]
\centering
\small
\caption{Per-iteration training accuracy on GSM8K-Hard (seed~42,
$N_{\text{train}}{=}30$). Values of 41.7\% require verification
(not achievable as integer/30; see Appendix note).}
\label{tab:iters}
\begin{tabular}{lcccc}
\toprule
\textbf{Method}
  & \textbf{Init} & \textbf{Iter~1} & \textbf{Iter~2}
  & \textbf{Iter~3} \\
\midrule
\gepa{}        & 33.3 & 36.7 & 40.0 & 40.0 \\
\mage-M        & 41.7 & 41.7 & 41.7 & 46.7 \\
\mage-P        & 33.3 & 41.7 & 41.7 & 26.7 \\
\mage-E        & 33.3 & 33.3 & 33.3 & 46.7 \\
\mage{} (full) & 41.7 & 50.0 & 50.0 & 53.3 \\
\bottomrule
\end{tabular}
\end{table}

\end{document}